\documentclass{article}

\usepackage[preprint, nonatbib]{neurips_2020}

\usepackage[utf8]{inputenc} % allow utf-8 input
\usepackage[T1]{fontenc}    % use 8-bit T1 fonts
\usepackage{hyperref}       % hyperlinks
\usepackage{url}            % simple URL typesetting
\usepackage{booktabs}       % professional-quality tables
\usepackage{amsfonts}       % blackboard math symbols
\usepackage{nicefrac}       % compact symbols for 1/2, etc.
\usepackage{microtype}      % microtypography
\usepackage[colorinlistoftodos]{todonotes} 
\usepackage{adjustbox}
\usepackage{graphicx}
\usepackage{tabularx}
\usepackage{xspace}

\def\tokenkind{{\em token\_kind }}
\def\cursorkind{{\em cursor\_kind }}

\newcommand{\cbert}{\texttt{C-BERT}\xspace}
\newcommand{\ffmpeg}{FFmpeg\xspace}
\newcommand{\qemu}{QEMU\xspace}
\newcommand{\langmodel}{LM\xspace}
\newcommand{\vulnerident}{VI\xspace}

\makeatletter
\newcommand{\printfnsymbol}[1]{%
  \textsuperscript{\@fnsymbol{#1}}%
}
\makeatother

\begin{document}

\title{Exploring Software Naturalness through\\Neural Language Models}

\author{Luca Buratti\thanks{Equal Contribution. In no particular order.} \\ IBM Research \\ \texttt{luca.buratti1@ibm.com}
\And Saurabh Pujar\printfnsymbol{1} \\ IBM Research \\ \texttt{saurabh.pujar@ibm.com}
\And Mihaela Bornea\printfnsymbol{1} \\ IBM Research \\ \texttt{mabornea@us.ibm.com}
\And Scott McCarley\printfnsymbol{1} \\ IBM Research \\ \texttt{jsmc@us.ibm.com}
\And Yunhui Zheng \\ IBM Research \\ \texttt{zhengyu@us.ibm.com}
\And Gaetano Rossiello \\ IBM Research \\ \texttt{gaetano.rossiello@ibm.com}
\And Alessandro Morari \\ IBM Research \\ \texttt{amorari@us.ibm.com}
\And Jim Laredo \\ IBM Research \\ \texttt{laredoj@us.ibm.com}
\And Veronika Thost \\ IBM Research \\ \texttt{veronika.thost@ibm.com}
\And Yufan Zhuang \\ IBM Research \\ \texttt{yufan.zhuang@ibm.com}
\And Giacomo Domeniconi \\ IBM Research \\ \texttt{giacomo.domeniconi1@ibm.com} \\
}

\maketitle

\begin{abstract}
The Software Naturalness hypothesis argues that programming languages can be understood through the same techniques used in natural language processing. We explore this hypothesis through the use of a pre-trained transformer-based language model to perform code analysis tasks. Present approaches to code analysis depend heavily on features derived from the Abstract Syntax Tree (AST) while our transformer-based language models work on raw source code. This work is the first to investigate whether such language models can discover AST features automatically. To achieve this, we introduce a sequence labeling task that directly probes the language model's understanding of AST. Our results show that transformer based language models achieve high accuracy in the AST tagging task. Furthermore, we evaluate our model on a software vulnerability identification task. Importantly, we show that our  approach obtains vulnerability identification results comparable to graph based approaches that rely heavily on compilers for feature extraction.

\end{abstract}

\section{Introduction} 
\label{sec:introduction}

The fields of Programming Languages (PL) and Natural Language Processing (NLP) have long relied on separate communities, approaches and techniques.
Researchers in the Software Engineering community have proposed the \textit{Software Naturalness}  hypothesis~\cite{naturalness}
which argues that programming languages can be understood and manipulated with the same approaches as natural languages. 

The idea of transferring representations, models and techniques from natural languages to programming languages has inspired interesting research. However, it raises  the question of whether language model approaches based purely on source code can compensate for the lack of structure and semantics available to graph-based approaches which incorporate compiler-produced features.
The application of language models to represent the source code has shown convincing results on code completion and bug detection~\cite{Hellendoorn2017,Karampatsis2020BigC}. 
The use of structured information, such as the Abstract Syntax Tree (AST), the Control Flow Graph (CFG) and the Data Flow Graph (DFG), has proven to be beneficial for vulnerability identification in source code~\cite{devign}. However, the extraction of structured information can be very costly, and requires the complete project. In the case of the C and C++ languages, this can only be done through complete pre-processing and compilation that includes all the libraries and source files. Because of these requirements, approaches based on structural information are not only computationally expensive but also inapplicable to incomplete code, for instance a pull request.

This work explores the software naturalness hypothesis, employing the pre-training/fine-tuning paradigm widely used with transformer-based~\cite{vaswani2017attention} language models (LMs)~\cite{bert,DBLP:conf/nips/WangPNSMHLB19}  to address tasks involving the analysis of both syntax and semantics of the source code in the C language.
To investigate syntax, we first introduce a novel sequence labeling task that directly probes the language model's understanding of AST, as produced by Clang \cite{csa}. 
Furthermore, we investigate the capabilities of LMs in handling complex semantics in the source code through the task of vulnerability identification (\vulnerident).
All of our experiments involved LMs pre-trained from scratch on the C language source code of 100 open source repositories.
The use of language models with source code rather than natural language leads to multiple challenges. 
Data sparsity is a major problem when building language models, leading to out-of-vocabulary (OOV) terms and poor representations for rare words. 
These issues are particularly severe for PL because variable and function names can be of almost arbitrary length and complexity.

There is a tradeoff between the granularity of tokenization and availability of long-range context for the LM.  Fine-grained tokenizations break identifiers into many small common tokens, alleviating issues with rare tokens, but at the risk of spreading important context across too many tokens.
We address the OOV and rare words problems by investigating three choices for tokenization,
which span the context/vocabulary size tradeoff 
from the extreme of character-based tokenization to subword tokenization styles familiar to the NLP community.
We indicate that the choice of the pre-training objective is closely connected to the choice of the vocabulary. In particular, with character based tokenization the pre-training task seems too easy. We introduce a more difficult, whole word masking (WWM) pre-training objective.

In our experiments, we show that our language model is able to effectively learn how to extract AST features from source code.
Moreover, we obtain compelling results compared to graph-based methods in the vulnerability identification task.
While current approaches to code analysis depend heavily on features derived from the AST~\cite{devign,Zhang2019ANov,Zhang2019ANov}, our approach works using raw source code without leveraging any kind of external features. This a major advantage, since it avoids a full compilation phase to avoid the extraction of structural information.
Indeed, our model can identify vulnerabilities during the software development stage or in artifacts with incomplete code, which is a valuable feature to increase productivity.
We show the merits of simple approaches to tokenization, obtaining the best results using the character based tokenization with WWM pre-training.

The contributions of this work are summarized as follows:
    1) we investigate the application of transformer-based language models for complex source code analysis tasks;
    2) we demonstrate that character-based tokenization and pre-training with WWM eliminate the OOV and rare words; 
    3) this work is the first to investigate whether such language models can discover AST features automatically;
    4) our language model outperforms graph-based methods that use the compiler to generate features.

\section{Model: \cbert}
\label{sec:bert_model}

We investigate the software naturalness hypothesis by pre-training from scratch a transformer-based LM, based on BERT~\cite{bert}, on a collection of repositories written in C language. 
Then, we fine-tune it on the AST node tagging (Section \ref{sec:ast_task}) and vulnerability identification (Section \ref{sec:va_task}) tasks.

However, the application of BERT on source code requires re-thinking the tokenization strategies.
Indeed, tokenizers based on language grammar are unsuitable for use with transformer language models because of the nearly unlimited resulting vocabulary size. (There are more than 4 million unique C tokens in our pre-training corpus.)
This greatly increases the training time of the language model, and introduces unacceptably many OOV items in held-out data if truncated to typical vocabulary sizes.
Encouraged by recent work on subword tokenization for source code \cite{Karampatsis2020BigC} we explore three subword tokenizer choices that reduce the vocabulary size and the impact of OOV and rare words.

\subsection{Tokenizers}
\label{sec:tokenizers}

\textbf{Character ($Char$)}
We investigate the extreme of subword tokenization: an ASCII vocabulary, an alternative dismissed by \cite{Karampatsis2020BigC}. 
Since our datasets are almost entirely ASCII characters, we are able to reduce the total vocabulary size to 103 (including [CLS], [SEP], [MASK] and [UNK].)
This choice minimizes vocabulary size, at the cost of limiting the size of context window available to the model.

\textbf{Character + Keyword ($KeyChar$)}
Most programming languages have a relatively small number of keywords.
By augmenting the $Char$ vocabulary with 32 C language keywords \cite{kandrC},
we are able to treat each keyword as a token, rather than breaking it up into individual characters or subwords.
This reduces the number of tokens per document (and increases the available context window), as can be seen in Table \ref{tab: token-stats}. 

\textbf{Sentencepiece ($SPE$)}
Neural NLP models almost always use a subword tokenizer such as Sentencepiece \cite{sentencepiece} or Byte-Pair Encoding (BPE) \cite{sennrich-etal-2016-neural}.
We follow other transformer based models and use a sentencepiece tokenizer, with vocabulary size chosen to be $5000$.
This includes all C keywords in the vocabulary, along with many common library functions and variable names.
We find that this leaves an almost-negligible number of tokens ($<0.001\%$) out-of-vocabulary in held-out data.
Table \ref{tab: token-stats} shows that $SPE$ tokenization reduces the number of tokens per document by about a factor of 3 compared to $Char$ and $KeyChar$.

\begin{table}
\caption{Average number of tokens per file in pre-training dataset. For comparison there were 3,272 C tokens per file in the pre-training data. Column 4 refers to the VI task (Section \ref{sec:va_task})}

\begin{center}
\begin{tabular}{|l|c|c|c|}
\hline
                   & \textbf{Vocab Size} & \textbf{Pre-training data} & \textbf{VI data} \\
\textbf{Tokenizer} & \textbf{(tokens)}             &   \textbf{(tokens/file)} & \textbf{(tokens/file)}\\
 \hline
$Char$ & 103 & 15,594 & 1789\\
$KeyChar$  & 135 & 15,011 & 1469 \\
$SPE$ & 5,000 & 5,500 & 558 \\
\hline
\end{tabular}
\end{center}
\label{tab: token-stats}
\end{table}

\subsection{Transformer Based Language Models. }
\label{sec:models}
Our model architecture is a multi-layer bidirectional transformer ~\cite{vaswani2017attention} based on BERT ~\cite{bert}, which we refer to as \cbert.
As in ~\cite{bert}, the language model component of our model is first pre-trained on a large unlabeled corpus.  
Task-specific training ("fine-tuning") is then continued with different task-specific objective functions on additional training data labeled for the tasks.
In this section, we briefly review the four objective functions and associated classifier layers involved in training our models.
%and we refer to our transformer language model as \cbert.
In all cases, our model is trained on fixed-width windows of tokens, to which we prepend a $[CLS]$ token to indicate the beginning of the sequence and appennd $[SEP]$ token to indicate the end of the sequence. 

For every input sequence $\mathbf{X} = [x_{1}=CLS, x_2,\ldots,x_{T-1},x_{T}=SEP]$, the language model outputs a sequence of contextualized token representations
$\mathbf{H} = [h_{1}=h_{CLS},h_2,  \ldots,h_{T}=h_{SEP}] \in \mathbb{R}^{T\times 768}$
where $h_t \in \mathbb{R}^{768}$.

\textbf{Masked Language Model (MLM) Pre-training Objective}
A small percentage of tokens are selected for "masking", as described in ~\cite{bert}.
A linear layer 
$\mathbf{W_{LM}} \in \mathbb{R}^{768\times |V|}$
($V$ denotes the \cbert vocabulary associated with a tokenization)
followed by \emph{softmax} is added on top of \cbert representation to compute the probability distribution over $V$ for each masked token.
The objective function is maximum likelihood of the true labels as computed from
\begin{equation}
p_{LM} = softmax(\mathbf{H} \mathbf{W_{LM}})
 \in \mathbb{R}^{768\times |V|}
\label{eq:linear}
\end{equation}
and evaluated over only the masked tokens.
The model trained with this objective is used to initialize the task-specific models.

\paragraph{Whole Word Masked (WWM) Pre-training Objective}
The masked language model task used to pretrain \cbert depends heavily on the tokenization.
Particularly with $Char$ tokenization, many masked tokens are ASCII characters within a variable name that is repeated elsewhere in the context.
We suspect in this case that the MLM task is too easy to adequately pretrain \cbert, and conjecture that making the pre-training task more difficult by masking longer spans of source code could result in stronger models on downstream tasks. 
Indeed, there is evidence that such techniques are beneficial in NLP. \cite{bertwwm,spanBERT}

As an alternative, we choose types of strings (for example, text strings that are legal variable or function names) to mask from the pre-training dataset, and write regular expressions to extract them. These matches are stored in a trie to create a dictionary. 
During pre-training, when a token is selected for masking, we identify which dictionary strings contain the token and then add the entire span of tokens from that string to the set of tokens masked in the MLM objective.  
We change the masking rate from 15\% in MLM, to 3\% for $Char$ and $KeyChar$ tokenizers and 9\% for $SPE$. The new probabilities ensure that the average number of masked tokens remains roughly the same as MLM.

\textbf{AST Fine-tuning Objective.} In this sequence labeling task, we add a linear layer $\mathbf{W_A} \in \mathbb{R}^{768\times |V_{AST}|}$
followed by \emph{softmax}.
Here $V_{AST}$ represents the set of AST labels.
We fine tune the model using the cross entropy between the gold AST labels and
\begin{equation}
p_{AST} = softmax(\mathbf{H} \mathbf{W_{AST}})
\label{eq:linear_ast}
\end{equation}
across all tokens.

\textbf{VI Fine-tuning Objective.} The VI is a binary classification task and depends only on 
the $h_{CLS}$ embedding.
We add a single linear layer $\mathbf{w} \in \mathbb{R}^{768}$.
We fine-tune using the cross entropy between the true labels and 
\begin{equation}
p_{VI} = sigmoid(h_{CLS}^{\top}  \mathbf{w})
\label{eq:linear_vi}
\end{equation}
evaluated across all context windows.

\subsection{Pre-training Corpus}

For the pre-training dataset we chose the top-100 starred (at least 5500 stars) GitHub C language repositories. Forks and clones were excluded to avoid duplication. These include well-maintained, widely-used repositories such as linux, git, postgres, and openssl.
The total size of the pre-training corpus is about 5.8 GB, much less than the corpus used to train BERT, or source code versions of BERT such as \cite{CuBERT} and \cite{CodeBert}. 
Comments were removed to keep only tokens related to code. We implemented the de-duplication strategy in \cite{deduplication} and found only about 40 files out of 60,000+ to be duplicates.

\subsection{Pre-training Details}
\label{sec:pretraindetails}
All of our models are built using the Huggingface Pytorch Transformers library~\cite{Wolf2019HuggingFacesTS}.
Regardless of the type of tokenizer used, our model consists of 12 layers of transformers, 768-dimensional embeddings, 12 attention heads/layers, the same architectural size of \textsc{BERTbase}~\cite{bert}.
We divide the training data into train, dev and test sets. Training is stopped based on dev set accuracy and models are selected for the VI task based on test set accuracy. We train on 8 nodes, with 32 GPUs, 512 maximum sequence length and batch size of 8, per GPU, for a total batch size of 256.
Learning rate and number of epochs vary based on the tokenizer. We found that $Char$ and $KeyChar$ models learn faster compared to the $SPE$ models. In total we selected 6 pretrained models, based on a combination of 3 tokenizers and 2 masking strategies.
We experiment with different learning rates (LR). 
$Char$ models with both MLM and WWM were trained with LR $10^{-4}$ and reach peak accuracy relatively quickly compared to the respective $KeyChar$ and $SPE$ models. For $KeyChar$ we used LR of $2 \times 10^{-5}$ for MLM and $10^{-4}$ for WWM. For $SPE$ we used LR of $2 \times 10^{-5}$ for MLM and $2 \times 10^{-6}$ for WWM.

\section{AST Node Tagging Task}
\label{sec:ast_task}
The Abstract Syntax Tree (AST) is an important structure produced by compilers and it has been used in prior works to improve the performance of code analysis tasks, such as vulnerability detection~\cite{devign}.
As a step toward full language model understanding of the AST, in this section, we describe a sequence labeling problem where the goal is to capture the syntactic role of each component of a linearized AST. Part-Of-Speech (POS) tagging is an analogous problem in NLP~\cite{manningTextbook}.

In detail, we propose two tasks to show that our models can discover the AST structure. For each task we represent the source code as a sequence of programming tokens, where each token is labeled with attributes \tokenkind and a \cursorkind. Gold labels for \tokenkind are produced by the compiler's tokenizer component, while gold labels for the \cursorkind are produced by the compiler's parser component which maps each token to its cursor node in the AST.  
The role of the language model is to predict the correct \tokenkind and \cursorkind for every programming token by examining the source code only.

To generate the gold data for our AST node tagging task we used Clang \cite{csa}, an open-source C/C++ compiler with a modular design. 
Clang\footnote{We use libclang python bindings atop Clang 11.0 } has $5$ \tokenkind labels and $209$ \cursorkind labels. 
The compiler handles the tokenization of the input source file and assigns the attribute  \tokenkind to each source code token.

The AST is represented as a directed acyclic graph of cursor nodes. The \cursorkind is an attribute of the cursor to indicate the type of the node in the AST. 
During parsing, every token in the source code is mapped to the corresponding cursor node in the AST according the rules of the C language grammar. 

The BERT-based language models and the baseline model solved the \tokenkind task with ease. 
Both accuracy and $F1$ were greater than $99\%$ across all three tokenizations on our data sets. 
On this basis we concluded that small differences in performance on \tokenkind tasks were unlikely 
to provide useful information about the strengths and weakness of our model. 
Further discussion will focus entirely on the \cursorkind task.

\begin{table}
\small
\caption{Example with $*$ used as multiplication (on the left); Example with $*$ used as pointer dereference (on the right)}
\begin{center}
\begin{adjustbox}{max width=\linewidth}
\begin{tabular}{|l|r|r||l|r|r|}

\hline
\textbf{token} & \textbf{\cursorkind} & \textbf{\tokenkind} & \textbf{token} & \textbf{\cursorkind} & \textbf{\tokenkind}\\
 \hline
if &     IF\_STMT & KEYWORD &sizeof	&  CXX\_UNARY\_EXPR & KEYWORD \\
( &      IF\_STMT & PUNCTUATION & ( &  PAREN\_EXPR & PUNCTUATION\\
lensum & DECL\_REF\_EXPR & IDENTIFIER & $*$     &  UNARY\_OPERATOR & PUNCTUATION\\
$*$ &      BINARY\_OPERATOR & PUNCTUATION & $*$ &  UNARY\_OPERATOR & PUNCTUATION\\
9 &      INTEGER\_LITERAL & LITERAL & s &  DECL\_REF\_EXPR & IDENTIFIER\\
/ &      BINARY\_OPERATOR & PUNCTUATION & ->&  MEMBER\_REF\_EXPR & PUNCTUATION\\
10 &     INTEGER\_LITERAL & LITERAL & coarse  &  MEMBER\_REF\_EXPR & IDENTIFIER\\
> &      BINARY\_OPERATOR & PUNCTUATION & )       &  PAREN\_EXPR & PUNCTUATION\\
maxpos & DECL\_REF\_EXPR & IDENTIFIER & & &\\
\hline

\end{tabular}
\end{adjustbox}
\end{center}
\label{clang-example-star-mult}
\end{table}

Predicting Clang's \cursorkind label is similar to the task of word sense disambiguation in NLP.
Just as the English word 'bank' may refer to a financial institution, or to the edge of a river,
many tokens produced by Clang have ambiguous meanings.  
For example the {\texttt *} operator is used as a binary operator, for multiplication,
and as a unary operator, for dereferencing a pointer variable (among other meanings.) 
We show examples of Clang output
for two C expressions containing {\texttt *} in Table \ref{clang-example-star-mult}.
In 
\begin{equation}
    \hbox{\texttt{ if ( lensum * 9 / 10 > maxpos )}}
\end{equation}
the {\texttt *} is used for multiplication, and is labeled BINARY\_OPERATOR by Clang.
In
\begin{equation}
    \hbox{\texttt{sizeof(**s->coarse)}}
\end{equation}
both instances of the {\texttt *} are used for pointer dereference, and are labeled UNARY \_OPERATOR.
The parentheses also have different labels in the two examples, and two meaning of the character {\texttt >} are implicitly distinguished
by its incorporation in the {\texttt ->} token.

\paragraph{Datasets} We created two annotated datasets using the code from the \ffmpeg~\cite{ffmpeg} and \qemu~\cite{qemu} GitHub repositories. 
\ffmpeg is a collection of libraries and tools to process multimedia content such as audio, video, subtitles and related metadata. It contains $1,751$ files with over 6 million Clang tokens. 
\qemu is a generic machine and  user space emulator and virtualizer containing $2,152$ source files and over 7 million Clang tokens. Both projects are open source.
We split each dataset by randomly assigning the files in a  $70$/$10$/$20$ ratio between train, dev and test sets.

We use the libclang python bindings atop Clang 11.0 to parse the C/C++ source files and traverse the AST. In order to get the precise context information (e.g. \cursorkind for tokens), we make sure the source files can be correctly compiled by intercepting the build process to obtain necessary compiler options and headers. 
We apply the LM tokenizer to the Clang output on each dataset and  retain a unique, deterministic alignment between the Clang tokens and the LM's tokens to produce predictions using \cbert, as explained in Section~\ref{sec:models}.

\paragraph{Fine-tuning Details} We initialize from pre-trained \cbert models for the three different tokenization strategies described in Section \ref{sec:bert_model} after validating that these models yielded acceptable performance on the VI task. The sequence classifier head in Equation~\ref{eq:linear_ast} was randomly initialized.

Selected experiments were repeated with 5 random number generator seeds in order to gauge fluctuations.
Since BERT is configured to handle a finite context window, the training file was divided into non-overlapping windows of maximum 250 tokens. For the dev and test data a sliding window approach is used to create chunks of 250 tokens with 32 tokens overlap.  When a token appears in multiple segments, its prediction is determined through voting. 
Models were trained using a batch size of 16, and for a maximum of 10 epochs or 24 hours, with final models selected on the basis of dev-set $F1$.
We investigated learning rates from the set of $\{5\times 10^{-5}, 2 \times 10^{-5}, 10^{-4}\}$   
For most experiments the learning rate of $2 \times 10^{-5}$ was used - we selected $10^{-4}$ 
if it yielded notably better accuracy during the learning rate exploration.
Most experiments were run on NVidia K-80s, with learning rate exploration on NVidia V-100s.

\paragraph{Experimental Results} Test set results for the \cursorkind tasks are shown in Table \ref{clang-results-cursor-test}.
We compare our transformer LMs with a BiLSTM baseline that uses the same tokenization as \cbert. Both systems produce \cursorkind predictions for every LM token in the input. We use the alignment between C tokens and LM tokens to report C token level F1 and accuracy for each dataset.  

Two trends are immediately apparent. 
First, BERT-based language models out-performed BiLSTM-based language models across all three tokenizations, and  on both the \ffmpeg and \qemu data sets. 
Second, performance on the \ffmpeg set was considerably higher than for the \qemu dataset, for both the BERT and the BiLSTM-based models.

Differences in tokenization style led to only small changes in performance for \cbert. 
These differences were not statistically significant - the absolute differences
between the min and max $F1$ across five initialization seeds  ranged from
$1\%$ to $3\%$ for the different tokenizations of \qemu, and ranged from $0.2\%$ to $0.4\%$ for the different tokenizations of \ffmpeg.
It appears that neural language models are able to infer enough higher-level structures of programming languages to perform these tasks without additional assistance from the tokenizer.
For the BiLSTM results the tokenization styles show no impact on the \ffmpeg dataset where both the accuracy and the F1 score are high. The tokenization has an impact on the \qemu dataset where the $SPE$ tokenizer performs the best.

We analyzed the confusion matrix for both \ffmpeg and \qemu dev sets.
In both, the top 2 errors were system predictions of COMPOUND\_STATEMENT or DECL\_REF\_EXPR when the reference indicated O (for outside of a \cursorkind).
Clang typically indicates O for \texttt{\#include} statements, macro definitions, and architecture-dependent conditional compilations that are skipped.
In particular, we note that both repositories contain long macro definitions which superficially resemble active code, but which require fairly long context to identify as a macro definition. The most common error not involving 'O' was predicting DECL\_REF\_EXPR for COMPOUND\_STMT.

\begin{table}
\caption{Test set accuracy and F1 for the AST \cursorkind tasks}
\begin{center}
\begin{adjustbox}{max width=\linewidth}
\begin{tabular}{|l||l||c|c||c|c|}
\hline
 &           &  \multicolumn{2}{c||}{\textbf{\ffmpeg}} &  \multicolumn{2}{c|}{\textbf{\qemu}} \\
\textbf{Model} & \textbf{Tokenizer} & \textbf{Acc}& \textbf{F1} & \textbf{Acc} & \textbf{F1} \\ 
\hline
 & $Char$  & 	     		94.96  &	95.71 &	71.68 &	80.53 \\
BiLSTM & $SPE$  & 		    	94.69  &	95.52 &	74.12  &	81.58 \\
 & $KeyChar$  &			95.68  &	96.58 &	66.20 &	76.19 \\
\hline										
 & $Char$  &		        	97.10   &	97.72 &	81.06 &	87.43 \\
\cbert & $SPE$  & 		        	97.72  &	98.29 &	\textbf{81.11} &	\textbf{87.79} \\
 & $KeyChar$  & 			\textbf{97.73}  &	\textbf{98.31} &	80.78 &	87.49 \\
\hline
\end{tabular}
\end{adjustbox}
\end{center}
\label{clang-results-cursor-test}
\end{table}

We also experimented with BERT models pre-trained with the whole-word masking objective.
These models hurt performance here, whereas (we will see in Section \ref{sec:va_task})  they improve VI performance.
For example, on the \qemu dev-set, $F1$ dropped from $93.3\%$ to $82.4\%$ for $Char$ tokenization, from $93.3\%$ to $88.2\%$ for $SPE$, and from $94.1\%$ to $90.8\%$ for $KeyChar$ tokenization. 
Further investigation is needed to determine whether this difference is due to the syntax/semantics focus of the two tasks or whether the WWM is simply more beneficial to a task such as VI with a lower baseline performance.

\section{Vulnerability Identification Task (VI)}
\label{sec:va_task}
While VI has been investigated previously (e.g. Draper\cite{draper}, Juliet\cite{juliet}, Devign\cite{devign}) there is no satisfactory comparable baseline.
The Draper dataset \cite{draper} has many false positives, and the Juliet dataset \cite{juliet} consists of synthetic data.
Devign \cite{devign} released appropriate natural data, but there are limitations which make it impossible to compare with their results. 
Only 2 out of 4 projects from the dataset were released, and the training/test split was not specified.
Also, unspecified data was omitted from their published results because of computational complexity and
limitations of Joern \cite{joern} preprocessing, a key step in their pipeline.
Furthermore, lack of details about a key feature of their GGNN \cite{ggnn} network, the pooling layer, prevent exact replication of the "Devign composite" model.

\paragraph{Data Description}
Our dataset consists of functions from the \ffmpeg and \qemu projects with non-vulnerable/vulnerable labels as released by Devign, with the duplicates removed. We also use these two datasets to create a combined dataset. The original \ffmpeg dataset contains 9,683 examples and \qemu  contains 17,515 examples. We call these three datasets  $full$ \ffmpeg, $full$ \qemu and $full$ combined datasets.
In order to implement the GGNN baseline, we use Joern to compute multiple graph features, including AST, just like Devign. 
We skip some examples that yield a compilation error with Joern.
With the exclusion of problematic samples, we get reduced versions of the full datasets, and we call the resulting datasets \ffmpeg $reduced$, \qemu $reduced$ and combined $reduced$.
The reduced dataset contains 6169 \ffmpeg and 14,896 \qemu examples, which is a total reduction of 6133 relatively large functions. 

\paragraph{Fine-tuning Details} 
We initialize with each of six pre-trained models described in Section \ref{sec:pretraindetails}.
The Huggingface implementation of BERT truncates function tokens beyond the count of 512. We call this default approach $Truncate$. We fine-tune according to Eq. \ref{eq:linear_vi}.
As can be seen in Table \ref{tab: token-stats}, column 4, using $Char$ and $KeyChar$ tokenizers often pushes the context token count beyond 512.
With $Char$ tokenizer, about 80\% of \ffmpeg functions have more than 510 tokens.
We aggregate such functions, similar to \cite{pappagari2019hierarchical},  by first breaking the input example into N different segments of maximum size 510 and then using a BiLSTM layer to combine the [CLS] output. 
This output is passed to the linear layer for classification.
We train for 10 Epochs, with learning rate of $2 \times 10^{-5}$ for $KeyChar$, $SPE$ and $8 \times 10^{-6}$ for $Char$, max sequence length of 512 and batch size of 4.

\paragraph{Baselines} The Naive baseline shows the accuracy if all instances are labeled vulnerable. 
Just like Devign, we use BiLSTM and CNN as baselines. Our strongest baseline, GGNN, is implemented as described in the Devign paper (GGRN composite) using the same graph features. 
Our implementation of the Devign composite model, which is GGNN with pooling, did not improve upon GGNN because we lacked adequate information on how to implement the pooling layer. We did not include Devign composite into our results.
We train BiLSTM and CNN baselines on all the six datasets. We  train GGNN only on the reduced datasets due to Joern compilation errors. Both BiLSTM and GGNN baselines are initialized with Word2Vec \cite{word2vec} embeddings trained on source code from the \ffmpeg and \qemu projects.

\begin{table}
\caption{Test set accuracy for the VI task on the  $full$ and $reduced$ ("$red$") datasets; \cbert model id indicates tokenization (C|K|S), LM masking objective (M|W) and aggregation(T|B); Aggr indicates the aggregation method}

\begin{center}
\begin{adjustbox}{max width=\linewidth}
\begin{tabular}{|l|l|l|l|c|c|c|c|c|c|}
\hline
     &           &           &           & \multicolumn{2}{c|}{\textbf{\ffmpeg}} & \multicolumn{2}{c|}{\textbf{\qemu}} &  \multicolumn{2}{c|}{\textbf{Combined}}  \\
\textbf{Model} & \textbf{LM + Tokenizer}& \textbf{Masking} & \textbf{Aggr}. & \textbf{\textit{full}} & \textbf{\textit{red}} & \textbf{\textit{full}} & \textbf{\textit{red}} & \textbf{\textit{full}} & \textbf{\textit{red}} \\ 
\hline
Naive &    &    &    & 51.1 & 46.5 & 42.4 & 41.1 & 45.5 & 42.7 \\
\hline
BiLSTM &    &    &    & 59.5 & 58.3 & 61.6 & 64.0 & 57.6 & 61.5 \\
CNN &    &    &    & 57.3 & 58.7 & 60.5 & 63.3 & 56.9 & 59.9 \\
GGNN &  &    &    & NA & 61.1 & NA & 65.8 & NA & 63.2 \\
\hline
\hline
CMT & \cbert $Char$ & MLM & Truncate & 52.7 & 54.7 & 57.3 & 58.7 & 55.5 & 57.8  \\
KMT & \cbert $KeyChar$ & MLM & Truncate & 55.5 & 57.0 & 57.5 & 59.5 & 56.2 & 58.1 \\
SMT & \cbert $SPE$ & MLM & Truncate & 57.7 & 54.8 & 59.3 & 60.5 & 57.4 & 58.0 \\
\hline										
CWB & \cbert $Char$ & WWM & BiLSTM & \textbf{62.2} & \textbf{65.5} & 65.8 & \textbf{68.1} & 63.5 & \textbf{66.2} \\
KWB & \cbert $KeyChar$ & WWM & BiLSTM & 58.0 & 61.9 & 64.1 & 67.7 & 61.5 & 64.3 \\
SMB & \cbert $SPE$ & MLM & BiLSTM & 60.7 & 62.8 & \textbf{66.1} & 66.4 & \textbf{63.6} & 65.4 \\
\hline
\end{tabular}
\end{adjustbox}
\end{center}

\label{tab:vi_results}
\end{table}

\paragraph{Experiment Results}
We report  results with accuracy as the evaluation metric rather than F1 score because our datasets are well-balanced, with 40\%-55\% labeled as vulnerable. The experiment results are in Table \ref{tab:vi_results}. 
\cbert models, with aggregation outperform the strongest GGNN baseline by a reasonable margin of 3-4 points across all datasets. They also perform better than BiLSTM and CNN baselines on both the full and reduced datasets.

The three lines CMT, KMT and SMT show the effect of varying the tokenization strategy when trained with the MLM objective. KMT performs better than CMT across all datasets and SMT has the best overall performance. As expected, tokenizations which enable the model to see longer context windows ($KeyChar$ and then $SPE$) achieve generally better results. All of our \cbert MLM models improve significantly upon the naive baseline. 

CWB, KWB and SMB are the best results for each of the three tokenizers.
The best model is CWB, showing the highest accuracy on most datasets. On the \qemu full and combined full, the CWB accuracy is comparable to SMB.
BiLSTM aggregation technique improves results across all datasets, for all models.
WWM improves the performance of $Char$ and $KeyChar$ tokenizer, as expected. It does not improve the $SPE$ tokenization. We found that WWM pre-training improves $Char$ the most, and mostly eliminates systematic differences in tokenization. Indeed, a $Char$ model, CWB, has the best results on 4 of the 6 datasets, while an $SPE$ model with MLM pre-training, SMB, is best on the other two.

\section{Related Work}
\label{related}
In recent years, research at the intersection of NLP, Machine Learning and Software Engineering has significantly increased. Topics in this field include
code completion~\cite{Raychev14codecompletion}, ~\cite{Hellendoorn2017},~\cite{Hussain2019DeepTL}, program repair~\cite{Chen2019Seq},~\cite{Santos2018Syntax},~\cite{Vasic2019NeuralPR}, bug detection~\cite{Ray2016Naturalness} and type inference~\cite{Raychev2015Pred},~\cite{Hellendoorn2018Deep}.~\cite{Malik2019NL2Type},~\cite{Pradel2019TypeWriter}. A common factor among all these techniques is the use of an n-gram or RNN based LM with the tokenization defined by the programming language. Thus they are severely affected by the OOV and rare words problems.\\ Sub-word tokenization (e.g. BPE) was first proposed by \cite{Karampatsis2020BigC} as a suitable alternative for source code and shows compelling results for code completion and bug detection. In our work, we show that finer grained tokenization techniques are beneficial for source code LMs compared to subword tokenizers.

CuBert \cite{CuBERT} is a recently introduced LM for modeling code. In contrast to our approach, this work does not consider character based tokenization and uses a subword tokenizer.
A Github corpus in Python is used for pre-training a BERT-like model which is fine-tuned for tasks like variable misuse classification, wrong binary operator detection, swapped operands, function-docstring mismatch, and prediction of exception type. 
CodeBert \cite{CodeBert} is another transformer LM specialized for code generation from textual descriptions. Their specialized pre-training uses bimodal examples of source code paired with natural language documentation from Github repositories.  

Prior to the use of machine learning techniques, static code analysis was used to understand code structures and apply rules handcrafted by programmers to identify potential vulnerabilities. Tools like RATS \cite{RATS}, Flawfinder \cite{Flawfinder}, and Infer \cite{Infer} are of this type.  However, they produce many false positives, a problem identified early on by \cite{Cheirdari_FP, Gadelha_CSA_FP, Johnson_SA_Survey}, making these tools difficult to use effectively as part of the developer tool chain. 

Current machine learning approaches to code analysis depend heavily on features derived from the compiler, such as the AST, or other derived structures that require the compilation of source code ~\cite{devign}. These features are paired with complex GGNN~\cite{msrggnn}  models. In addition to the VI task ~\cite{devign}, combining the AST and GGNN have shown good results in type inference~\cite{Wei2020LambdaNet}, code clone detection~\cite{Chen2019Capt, Zhang2019Novel} and bug detection~\cite{Liang2019DeepLW}. This evidence motivated our AST tagging task.

\section{Conclusions}
\label{sec:conclusions}
This work explores the  software naturalness hypothesis by using language models on multiple source code tasks.
Unlike graph based approaches that use structural features like the AST, CFG and DFG, our model is built on raw source code. Furthermore, we show that the AST \tokenkind and \cursorkind can be learned with the LM. LMs can work even better than graph based approaches for VI because they avoid the computational cost and requirements of full compilation. This makes our approach suitable for a wider range of scenarios, such as pull requests.
We propose two character based tokenization approaches that solve the OOV problem while having very small vocabularies. We suggest ways to improve them with aggregation and WWM. These approaches work just as well and sometimes even better than a subword tokenizer like sentencepiece that has been previously explored for source code. 
In future work we propose joint learning of the AST and VI tasks on top of LM to further improve code analysis, without using the compiler to extract structured information.

\section*{Broader Impact}

Our research supports the software naturalness hypothesis. This means that it may be possible to transfer many recent advances in natural language understanding into the software domain and improve source code understanding. The \langmodel created by pre-training on source code can thus be used for different tasks such as \vulnerident, code completion \cite{Raychev14codecompletion}, code repair \cite{Vasic2019NeuralPR}, as well as multi-modal tasks involving both natural language and source code \cite{CodeBert}. These tasks have applications in the fields of software security, developer tools and automation of software development and maintenance. Here we focus on  software security through the task of VI.
Development and improvement of end-to-end systems that can identify software vulnerabilities will make it easier to do the same in open source software. As these tools become more available to everyone, it becomes imperative that developers protect their code from malicious agents by incorporating these tools in Continuous Integration/Continuous Delivery pipelines to identify vulnerabilities before code is exposed to others.

\bibliographystyle{acm}
\bibliography{bibliography}

\end{document}